\documentclass[letterpaper, 10 pt, conference]{ieeeconf}  % Comment this line out if you need a4paper
\usepackage{amssymb}
\usepackage{amsmath}
\usepackage{graphicx} % Add this line to include the graphicx package
\usepackage{multirow}
\usepackage{newtxmath}
\usepackage{cite}    
\makeatletter
\let\NAT@parse\undefined
\makeatother
\usepackage[colorlinks=true,linkcolor=blue,citecolor=blue,urlcolor=blue,pdftex]{hyperref}
\usepackage{algorithm}
\usepackage{algorithmic}
\usepackage{tabularx}
\usepackage{booktabs}

\IEEEoverridecommandlockouts                              % This command is only needed if 
\title{\LARGE \bf
Learning Motion Skills with Adaptive Assistive Curriculum Force in Humanoid Robots
}

\author{Zhanxiang Cao, Yang Zhang, Buqing Nie, Huangxuan Lin, Haoyang Li, and Yue Gao$^{\dag}$}% <-this % stops a space
\begin{document}

\maketitle
\thispagestyle{empty}
\pagestyle{empty}

%%%%%%%%%%%%%%%%%%%%%%%%%%%%%%%%%%%%%%%%%%%%%%%%%%%%%%%%%%%%%%%%%%%%%%%%%%%%%%%%
\begin{abstract}
    Learning policies for complex humanoid tasks remains both challenging and compelling. Inspired by how infants and athletes rely on external support—such as parental walkers or coach-applied guidance—to acquire skills like walking, dancing, and performing acrobatic flips, we propose \textbf{A2CF}: \textit{Adaptive Assistive Curriculum Force} for humanoid motion learning. A2CF trains a dual-agent system, in which a dedicated \textit{assistive force agent} applies state-dependent forces to guide the robot through difficult initial motions and gradually reduces assistance as the robot's proficiency improves. Across three benchmarks—bipedal walking, choreographed dancing, and backflips—A2CF achieves convergence 30\% faster than baseline methods, lowers failure rates by over 40\%, and ultimately produces robust, support-free policies. Real-world experiments further demonstrate that adaptively applied assistive forces significantly accelerate the acquisition of complex skills in high-dimensional robotic control.
\end{abstract}

%%%%%%%%%%%%%%%%%%%%%%%%%%%%%%%%%%%%%%%%%%%%%%%%%%%%%%%%%%%%%%%%%%%%%%%%%%%%%%%%
\section{INTRODUCTION}
The development of humanoid robots capable of learning complex motion skills, such as dancing and acrobatic movements, remains a significant challenge~\cite{tong2024advancements, gu2025humanoid}. Despite recent advancements in reinforcement learning (RL)\cite{radosavovic2024real, xue2025unified, wang2025beamdojo} and imitation learning (IL)\cite{he2025asap, he2024omnih2o, ji2024exbody2, liu2024opt2skill}, robots still struggle to acquire such skills effectively, particularly in terms of the stability and efficiency of the learning process. A key challenge in this domain is the balance between exploration and exploitation, which often results in slow learning and suboptimal performance~\cite{kwa2022balancing, ladosz2022exploration}. These limitations highlight the need for more effective learning strategies that can improve both the speed and performance of skill acquisition, especially for high-dimensional humanoid control tasks.
% To address these issues, this work proposes a reinforcement learning-based motion control framework that integrates assistive forces. This approach is designed to accelerate skill acquisition and enhance the performance of learned behaviors.

During human development, external assistance plays a crucial role in learning motion skills~\cite{wulf2007attention}. Infants, for example, often rely on parental support during their first steps, with walkers or direct physical assistance to help them gain the confidence and balance needed for independent locomotion~\cite{claxton2012control, von1982eye}. Similarly, in the case of highly complex movements like backflips, experienced coaches provide physical guidance, supporting the learner's back and applying upward forces to prevent falls and promote proper technique~\cite{werner2012teaching}. Studies indicate that such external aids not only expedite the learning process but also help prevent learners from adopting ineffective or unsafe strategies~\cite{dowdell2010characteristics}. The assistance prevents learners from getting trapped in local optima, enabling them to discover more efficient and stable movement strategies. This observation serves as a basis for our proposed framework, which incorporates assistive forces during robotic learning.

\begin{figure}[htbp]
    \centering
    \includegraphics[width=1.0\linewidth]{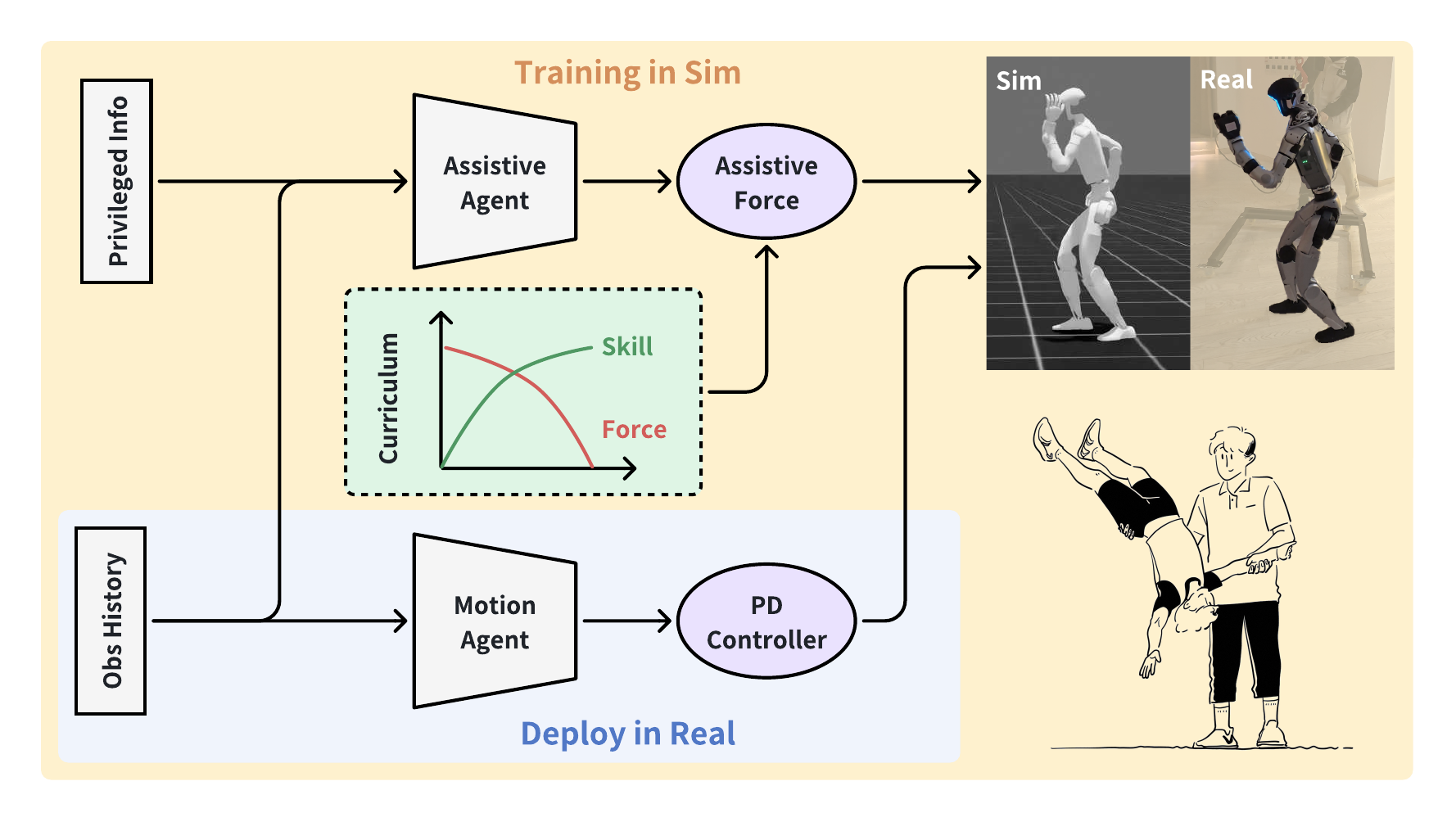}
    \caption{\textbf{Overall Algorithm Framework.} The figure illustrates the pipeline of the algorithm. In the simulation, the assistive force agent applies assistive forces to accelerate learning. With the help of the force bound curriculum, the assistive force is gradually optimized to zero. In the real-world deployment, no assistive force is required.}
    \label{fig:pipeline}
\end{figure}

However, robotic systems face unique challenges in acquiring complex motion strategies. Early-stage learning often leads to inefficient exploration, where the robot may fail to discover effective motions or get trapped in suboptimal solutions~\cite{tong2024advancements}. Unlike humans, robots are generally unable to rely on instinctive physical support during training, making the learning process more prone to instability. Furthermore, the difficulty of maintaining balance and ensuring robustness further complicates the acquisition of sophisticated movements. To address these issues, we draw inspiration from human learning by incorporating assistive forces into the reinforcement learning framework. These forces assist the robot in the early phases of skill acquisition, enabling it to learn more effectively and with greater stability.

% \subsection{Contributions}
In this paper, we propose a reinforcement learning framework that integrates adaptive assistive forces to accelerate the acquisition of humanoid robot motion skills. Initially, strong assistive forces guide the robot to facilitate efficient learning; these forces are gradually reduced in accordance with the robot's progress until fully withdrawn. This strategy emulates human learning support and enables the robot to ultimately perform tasks independently. We validate the approach on three tasks—walking, dancing, and backflips—demonstrating that adaptive assistance significantly enhances both learning efficiency and final performance in real-world deployments.

Our primary contributions are as follows:
\begin{enumerate}
    \item We introduce \textbf{A2CF}, a RL framework incorporating \textbf{A}daptive \textbf{A}ssistive \textbf{C}urriculum \textbf{F}orce to accelerate the learning of complex humanoid motions. An assistive force agent is jointly trained with the motion agent to provide state-dependent guidance.
    \item We enhance the assistive learning paradigm by integrating privileged information, tailored initial state distributions, and random masking, inspired by human motor learning, to improve generalization and prevent over-reliance on external support.
    \item We conduct both simulation and real-world experiments, demonstrating that A2CF yields notable improvements in learning speed and task performance, and successfully transfers to physical humanoid robots.
\end{enumerate}

\section{RELATED WORKS}

\subsection{Learning-Based Locomotion in Humanoid}

Recent advancements in learning-based methods have significantly enhanced humanoid locomotion. Radosavovic et al.~\cite{radosavovic2024real} proposed a transformer-based RL policy for zero-shot deployment on real robots, enabling robust walking across various terrains. HiLo~\cite{zhang2025hilo} developed RL policies that combine motion tracking with domain randomization to achieve natural, resilient movement. Additionally, Gu et al.~\cite{gu2024advancing} introduced the DWL framework, which allows humanoid robots to master complex terrains like snow and stairs through zero-shot simulation-to-reality transfer, demonstrating high robustness and generalization.

Furthermore, HoST~\cite{huang2025learning} and HumanUp~\cite{he2025learning} focus on training humanoid robots to recover from postures such as lying down and crawling to standing. In HoST, an externally applied vertical force accelerates the learning of the recovery motion, but this force is not adaptive and is independent of the robot's state.

\subsection{Imitation Learning for Humanoid Whole-Body Control}

Imitation learning has become a powerful method for teaching humanoid robots complex whole-body movements by mimicking human demonstrations. Zhang et al.~\cite{zhang2024whole} applied AMP~\cite{peng2021amp} to humanoid robots, introducing human walking trajectory priors to make the robot's gait more human-like. Similarly, OmniH2O~\cite{he2024omnih2o} and ExBody2~\cite{ji2024exbody2} achieve dexterous whole-body imitation by retargeting human motion trajectories to specific robots.

The TRILL system~\cite{seo2023deep} utilizes deep imitation learning for loco-manipulation tasks, while Matsuura et al.~\cite{matsuura2023development} developed a whole-body imitation learning system for a biped and bi-armed humanoid robot. Both systems integrate teleoperation and deep learning to enable robots to perform complex tasks like fabric manipulation and heavy lifting. These studies collectively demonstrate the effectiveness of imitation learning in enabling humanoid robots to perform human-like movements, bridging the gap between human demonstrations and robotic execution.

\section{METHODOLOGY}

\subsection{Problem Statement}
In this work, we model the task of learning motion skills as a Partially Observable Markov Decision Process (POMDP), denoted by the tuple \( M = (\mathcal{S}, \mathcal{A}, \mathcal{T}, \mathcal{R}, \Omega, \mathcal{O}, \gamma) \). Specifically, \( \mathcal{S} \) denotes the state space, which consists of the set of states \( s_t \) at each time step \( t \). The state includes the observation \( o_t \), as well as additional privileged information \( o_t^{priv} \) available during simulation. The observation \( o_t \) includes sensory data that the robot can access, and it is expressed as:
\[
    o_t = [w_t, g_t, q_t, \dot{q}_t, a_{t-1}, c_t],
\]
where \( w_t \) represents the angular velocity, \( g_t \) is the gravity projection vector of the pose, \( q_t \) and \( \dot{q}_t \) denote the joint positions and velocities, respectively, \( a_{t-1} \) is the previous action, and \( c_t \) is the current command. The observation space \( \Omega \) is the set of all possible observations \( o_t \).

The action space \( \mathcal{A} \) is the set of actions \( a_t \) that the agent can execute. The environment evolves according to the state transition function \( \mathcal{T}(s_{t+1} | s_t, a_t) \), which defines the probability distribution of the next state \( s_{t+1} \) given the current state \( s_t \) and action \( a_t \). Upon transitioning to a new state, the agent receives a reward \( r_t \) from the reward function \( \mathcal{R}(s_t, a_t) \) and observes the new state \( o_t \in \Omega \) via the observation function \( \mathcal{O}(o_t | s_{t+1}, a_t) \).

The goal of the agent is to learn an optimal policy \( \pi^* \) that maximizes the expected cumulative reward \( J(\pi) \) over an infinite time horizon. This is formalized as:
\[
    J(\pi) = \mathbb{E}_{\pi} \left[ \sum_{t=0}^{\infty} \gamma^t \mathcal{R}(s_t, a_t) \right],
\]
where \( \gamma \in [0, 1) \) is the discount factor that determines the weight given to future rewards.

% \begin{figure}[htbp]
%     \centering
%     \includegraphics[width=1.0\linewidth]{figures/pipeline.pdf}
%     \caption{\textbf{Overall Algorithm Pipeline.} The figure illustrates the pipeline of the algorithm. In the simulation, the assistive force agent provides assistive forces to accelerate learning. With the help of the force limit curriculum, the assistive force is gradually optimized to zero. In the real-world deployment, no assistive force is required.}
%     \label{fig:pipeline}
% \end{figure}

\subsection{Learning with Adaptive Assistive Spatial Force (\textbf{A2CF})}

\subsubsection{Expanded Action Space}
In addition to training the motion policy agent, an assistive force agent is trained to provide additional assistance to the robot during the training process. The close collaboration between the two agents is facilitated by joint action learners (JAL)~\cite{  canese2021multi, du2023review}. The action \( a_t^{\text{motion}} \) of the motion policy agent corresponds to the desired joint position \( q_t^{\text{des}} \), which is subsequently controlled by a PD controller. Based on this, the action space for the assistive force agent, \( a_t^{\text{assi}} \), is expanded, where the 6-D spatial force \( F_t = [f_t, m_t]  \), with \( f_t \) representing the linear force and \( m_t \) representing the moment. For convenience, this force is applied to the robot's base link, specifically the pelvis of the humanoid robot. Thus, the expanded action space becomes \( a_t = [a_t^{\text{motion}}, a_t^{\text{assi}}] \), which accelerates the learning process and enhances performance by simultaneously adjusting both the joint positions and the assistive forces during training.

\subsubsection{Assistive Force Curriculum}

In human motor learning—such as walking, backflipping, or dancing—external assistance from parents or coaches is typically provided during the early learning stages and gradually removed as the learner gains proficiency. Inspired by this observation, we introduce a curriculum-based mechanism for regulating assistive forces in robotic motion learning. Specifically, we define a bounded action space for the assistive force agent as a 6D hypercube centered at the origin:
\[
    \mathcal{B}_k = \left\{ F \in \mathbb{R}^6 \mid -\eta_{k, i} \leq F_i \leq \eta_{k, i}, \quad \forall i \in \{1,\dots,6\} \right\},
\]
where \( \eta_k \in \mathbb{R}^+ \) denotes the half-width of the hypercube at training iteration \( k \), and each dimension corresponds to a linear or torque component of the spatial assistive force.

To ensure a gradual reduction in assistance, the bound \( \eta_k \) is adaptively updated based on the magnitude of the applied force \( F_k \) and a skill acquisition indicator. The update rule is described in Algorithm~\ref{alg:curriculum}. When the normalized magnitude \( \|F_k\| / \| \eta_k \| \) falls below a threshold, the system infers that the assistive contribution is diminishing and decreases \( \eta_k \) accordingly. If the motion policy has completed skill acquisition, a decay is also enforced to eliminate external support during deployment.

\begin{algorithm}[htbp]
    \caption{Hypercube-Based Assistive Force Curriculum}
    \label{alg:curriculum}
    \begin{algorithmic}[1]
        \STATE \textbf{Input:} Initial bound \( \eta_0 \), applied force \( F_k \), skill acquisition flag \texttt{isSkillLearned}
        \STATE \textbf{Output:} Updated bound \( \eta_k \)
        \FOR{each training iteration \( k = 1, 2, \dots \)}
        \IF{\( \|F_k\| < (1 - \epsilon) \cdot \| \eta_k \| \)}
        \STATE \( \eta_k \gets (1 - \delta) \cdot \eta_k \)
        \ELSIF{\( \|F_k\| > (1 + \epsilon) \cdot \| \eta_k \| \)}
        \STATE \( \eta_k \gets (1 + \delta) \cdot \eta_k \)
        \ENDIF
        \IF{\texttt{isSkillLearned}}
        \STATE \( \eta_k \gets (1 - \delta) \cdot \eta_k \)
        \ENDIF
        \ENDFOR
    \end{algorithmic}
\end{algorithm}

\subsubsection{Initial Distribution of Spatial Force (\textbf{ID})}\label{ID}

An appropriately designed initial bound of the assistive force hypercube can provide the assistive force agent with a strong prior, thereby improving learning efficiency and stability. For example, in the training of walking skills, assistive forces are predominantly required in the lateral \(xy\)-plane, while the demand along the vertical \(z\)-axis is minimal. In contrast, more dynamic tasks such as backflips exhibit phase-dependent variations in assistive force requirements: during the \textit{Stand} and \textit{Land} phases, lateral assistance in the \(xy\)-plane is more critical, whereas the \textit{Jump} and \textit{Air} phases necessitate stronger support in the \(z\)-direction to generate and control vertical momentum.

To encode such human-inspired priors into the learning process, we initialize the spatial force bound \(\eta_0\) in a task-dependent manner. These initial bounds guide the early behavior of the assistive force agent by shaping the feasible region of the applied forces. The specific initialization bounds for each task are described in detail in Section~\ref{setup}.

% The robot agent receives both the current observation \( o_t \) and the historical information \( o_t^H \), and generates the desired joint positions as its output. In contrast, the assistive force agent processes the same inputs, \( o_t \) and \( o_t^H \), but is additionally provided with privileged information \( o_t^{priv} \) from the simulation environment. This extended input enables the assistive force agent to output a six-dimensional spatial force vector \( F_t = [F^{linear}_t, M^{torque}_t] \), where \( F^{linear}_t \) denotes the applied force and \( M^{torque}_t \) represents the torque exerted on the robot's base link. To capture the implicit information contained within the historical observations \( o_t^H \), we employ a Variational Autoencoder (VAE) network, as proposed in \cite{nahrendra2023dreamwaq}. This allows us to efficiently encode the historical data. To ensure effective coordination, both agents share the same portion of the embedded vector \( z_t \) derived from \( o_t^H \), thus enabling synergistic interaction while preserving their individual functionalities.

\subsubsection{Assistive Force Agent with Privileged Information (\textbf{PI})}\label{PI}

Since the assistive force agent only operates in simulation, it can fully utilize privileged information available in the simulation, as demonstrated by Lee et al.~\cite{lee2020learning}. During the training of walking skills, the assistive force agent can leverage terrain information to provide assistive forces that align with the current terrain conditions. This enables the motion policy agent to overcome difficulties more efficiently, helping it escape potentially conservative local optima and more quickly navigate challenging terrains.

In addition to terrain information, privileged information such as the current assistive force bound, robot linear velocity, and domain randomization parameters (e.g., ground friction coefficient, disturbance speed, and mass) are also used. This is similar to how a parent might assist a child in learning to walk by observing many details that the child cannot perceive themselves, offering additional support to facilitate faster learning and progression.

% \subsection{Initial Distribution of Spatial Force (\textbf{ID})}\label{ID}

% A good initial distribution can provide the assistive force agent with a solid prior. For instance, during the training of walking skills, the primary need is often lateral force in the \(xy\)-plane, while the need for longitudinal force along the \(z\)-axis is relatively small. However, for more complex tasks such as backflip, the demand for assistive force varies across different phases. In the Stand and Land phases, the need for lateral force in the \(xy\)-direction is greater, whereas in the Jump and Air phases, the requirement for longitudinal force along the \(z\)-axis increases. We incorporate human priors by providing different initial bounds for the \( F_t \) at the beginning of the task. The specific initial bounds are detailed in Section~\ref{setup}.

\subsubsection{Random Mask of Spatial Force (\textbf{RM})}\label{RM}
% TODO
During the training process, when assistive forces are large and persistent, the motion policy agent may become overly dependent on these forces, neglecting to execute tasks through its own joint control. To mitigate this dependency, we introduce a random masking mechanism during the application of assistive forces. This mechanism ensures that assistive forces are not always present, thereby encouraging the agent to gradually learn how to perform tasks independently.

Specifically, during each training iteration, a random mask is applied to selectively disable certain components of the assistive force with a probability of \(\zeta \). The final assistive force \( F^\text{assi}_t \) applied to the robot is expressed as:
\[
    F^\text{assi}_t = M_t \odot F_t
\]
where \( M_t \in \{0, 1\} \) is a randomly generated mask, with \( P(M_t = 0) = \zeta \) and \( P(M_t = 1) = 1 - \zeta \), and \( F_t \) represents the output of the assistive force agent. By employing this method, the robot is occasionally required to complete tasks without the aid of assistive forces, similar to how a parent occasionally lets go when teaching a child to walk, allowing the child to develop the ability to walk independently.

This strategy effectively reduces the agent’s reliance on assistive forces and accelerates the development of autonomous motion control skills, ultimately enabling efficient task execution without the need for assistive forces.

\subsection{Overall Training Architecture}

For the POMDP problem, an Asymmetric Actor-Critic (AAC) structure \cite{pinto2017asymmetric} is employed for training. The motion policy agent receives only the observation \( o_t \) and historical information \( o_t^H \), while the Critic receives the full state \( s_t \), which includes both the observation \( o_t \) and privileged information \( o_t^{\text{priv}} \), which is only available in simulation. To better leverage the historical information \( o_t^H \), the architecture proposed in \cite{nahrendra2023dreamwaq, zhang2024robust} is adopted, utilizing a Variational Autoencoder (VAE) \cite{higgins2017beta, burgess2018understanding} to compress the historical data and estimate the robot's linear velocity \( v_t \). The VAE takes in the historical information \( o_t^H \) and attempts to predict the next observation \( o_{t+1} \). The optimization is performed by minimizing the MSE loss between the predicted and real values of \( o_{t+1} \), as well as the KL divergence between the latent space and a standard normal distribution \( \mathcal{N}(0, I) \).

Both the motion policy agent and the assistive force agent share the compressed latent representation \( z_t \) from the historical information, along with the estimated velocity \( v_t \) and the observation \( o_t \) as inputs. Additionally, the assistive force agent receives a subset of the privileged information as input, which is explained in detail in Section~\ref{PI}. The PPO algorithm \cite{schulman2017proximal} is used to optimize the policy.

\subsection{Task-Specific Training Frameworks}
This section describes the task-specific training pipelines and reward designs for three locomotion tasks. The overall reward function consists of two components: the motion reward \( r^{\text{motion}} \) and the assistive force reward \( r^{\text{force}} \). The former encourages the acquisition of desired motion skills and includes both task-relevant terms and regularization terms, while the latter penalizes the excessive use of assistive forces to ensure minimal reliance on external support. The assistive force reward \( r^{\text{force}} \) is only introduced once the motion agent achieves 80\% of the skill acquisition target.

\subsubsection{Walking}
In the walking task, a locomotion policy is trained to follow a command velocity vector \( \mathbf{c}_t = [v_x^{\text{cmd}}, v_y^{\text{cmd}}, \omega_z^{\text{cmd}}] \), representing the desired forward and lateral linear velocities and the angular velocity of the yaw, respectively. The task reward is primarily based on the exponential penalty of velocity tracking error. Additionally, a cost of transport term, adapted from \cite{fu2021minimizing}, is introduced to optimize the energy efficiency of the gait. The detailed reward components are summarized in Table~\ref{tab:walk_rew}.

\begin{table}[htbp]
    \centering
    \caption{Reward Terms for Walking Task.}
    \label{tab:walk_rew}
    \begin{tabular}{llll}
        \toprule
        \textbf{Type} & \textbf{Name}     & \textbf{Equation}                                                         & \textbf{Weight} \\
        \midrule
        \multirow{2}{*}{Task}
                      & Lin. Vel. Track   & \( \exp[-4(v_{xy} - v_{xy}^{\text{cmd}})^2] \)                            & 5.0             \\
                      & Ang. Vel. Track   & \( \exp[-4(\omega_z - \omega_z^{\text{cmd}})^2] \)                        & 2.0             \\
        \midrule
        \multirow{8}{*}{Reg.}
                      & Feet Air Time     & \( \sum_i \vmathbb{1}_{\text{air}, i} \cdot t_{\text{air}, i} \)          & 1.0             \\
                      & Feet Slide        & \( \sum_i \vmathbb{1}_{\text{contact}, i} \cdot \|v_{foot, i}\| \)        & -1.0            \\
                      & Cost of Transport & \( \frac{\|\dot{q}\|\|\tau\|}{9.81\|\mathbf{v}_{xy}\|} \)                 & -0.01           \\
                      & Stand Still       & \( \vmathbb{1}_{\|\mathbf{c}\| < 0.1} \cdot \|q - q_{\text{default}}\| \) & -0.1            \\
                      & Action Rate       & \( \|a_t - a_{t-1}\|^2 \)                                                 & -0.01           \\
                      & DOF Acc.          & \( \|\ddot{q}\|^2 \)                                                      & -2.5e-7         \\
                      & DOF Pos. Limit    & \(\sum_j (\| q_j \| - q_j^\text{lim})\)                                   & -1.0            \\
                      & Orientation       & \( \|g\| \)                                                               & -1.0            \\
        \midrule
        Force         & Less Assi. Force  & \( \exp[-2\|F\| / \| \eta\|] \)                                           & 2.0             \\
        \bottomrule
    \end{tabular}
\end{table}

\subsubsection{Backflip}
\begin{figure}[htbp]
    \centering
    \includegraphics[width=1.0\linewidth]{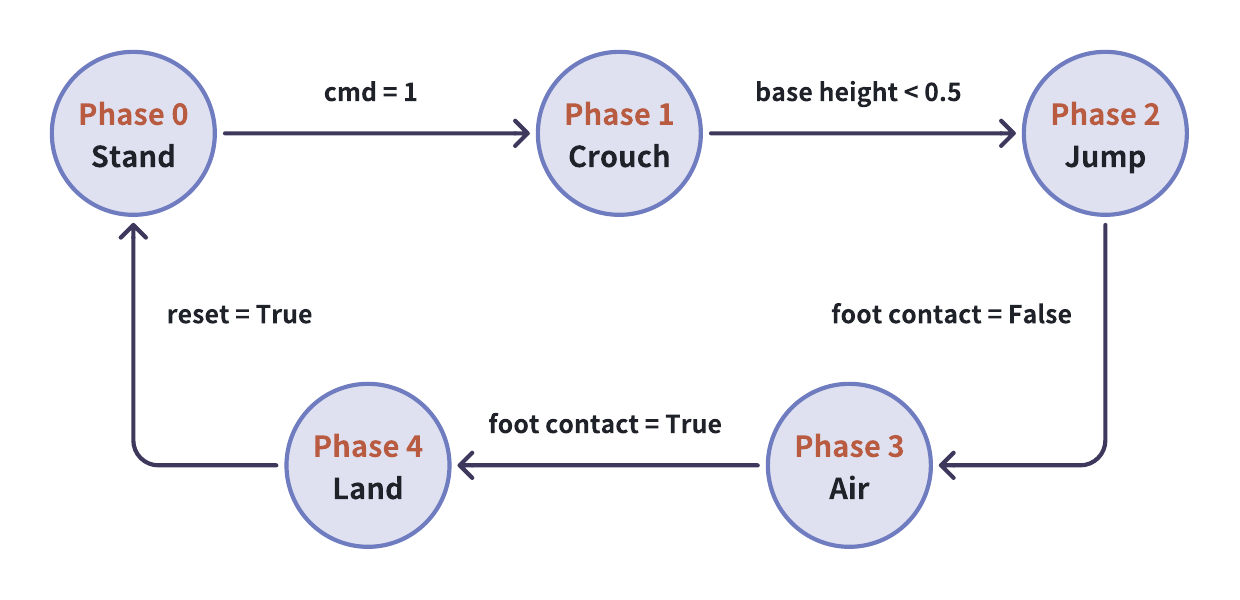}
    \caption{\textbf{FSM for Backflip.} The figure shows the five states of the backflip task, with arrows indicating state transitions. The conditions for these transitions are shown along the horizontal lines.}
    \label{fig:FSM}
\end{figure}

For complex motion tasks such as backflip, a phase-based training approach is adopted following the method in \cite{kim2024stage}. The environment maintains a finite state machine (FSM) with five discrete phases: \textit{Stand}, \textit{Crouch}, \textit{Jump}, \textit{Air}, and \textit{Land}, as illustrated in Fig.~\ref{fig:FSM}. Transitions between phases are primarily determined by the robot's height and foot contact states.

The task command \( \mathbf{c}_t \) is defined as a simple start trigger \( \mathbf{c}_t = \vmathbb{1}_{\text{start}} \), which also acts as the condition for transitioning from the \textit{Stand} phase to \textit{Crouch}. Based on the current phase, the reward function is redesigned to guide learning at each phase. In particular, additional vertical velocity rewards are introduced during the \textit{Crouch} and \textit{Jump} phases to encourage rapid body movement. The complete reward design is summarized in Table~\ref{tab:backflip_rew}.

\begin{table}[htbp]
    \centering
    \caption{Reward Terms for Backflip Task.}
    \label{tab:backflip_rew}
    \begin{tabular}{llll}
        \toprule
        \textbf{Type} & \textbf{Name}        & \textbf{Equation}                                                                                    & \textbf{Weight} \\
        \midrule
        \multirow{6}{*}{Task}
                      & Alive (Phase 0)      & \( \vmathbb{1}_{\text{phase}_0} \)                                                                   & 2.0             \\
                      & Down. Vel. (Phase 1) & \( \vmathbb{1}_{\text{phase}_1} \cdot \vmathbb{1}_{\text{contact}} \cdot -v_z \)                     & 2.0             \\
                      & Up. Vel. (Phase 2)   & \( \vmathbb{1}_{\text{phase}_2} \cdot v_z \)                                                         & 2.0             \\
                      & Ang. Vel. (Phase 2)  & \( \vmathbb{1}_{\text{phase}_2} \cdot -w_y \)                                                        & 0.5             \\
                      & Ang. Vel. (Phase 3)  & \( \vmathbb{1}_{\text{phase}_3} \cdot -w_y \)                                                        & 2.0             \\
                      & Land (Phase 4)       & \( \vmathbb{1}_{\text{phase}_4} \cdot exp[-10\|v\|] \)                                               & 20.0            \\
        \midrule
        \multirow{7}{*}{Reg.}
                      & Balance(Phase 0,1,4) & \( \vmathbb{1}_{\text{phase}_{0,1,4}} \cdot \angle \mathbf{z}_\text{base}, \mathbf{z}_\text{world}\) & -2.0            \\
                      & Balance(Phase 2,3)   & \( \vmathbb{1}_{\text{phase}_{2,3}} \cdot \angle \mathbf{y}_\text{base}, \mathbf{y}_\text{world}\)   & -2.0            \\
                      & Yaw Ang. Vel.        & \( \| w_z \| \)                                                                                      & -0.1            \\
                      & Action Rate          & \( \|a_t - a_{t-1}\|^2 \)                                                                            & -0.01           \\
                      & Joint Acc.           & \( \|\ddot{q}\|^2 \)                                                                                 & -2.5e-7         \\
                      & DOF Pos. Limit       & \(\sum_j (\| q_j \| - q_j^\text{lim})\)                                                              & -1.0            \\
                      & DOF Pos. Deviation   & \( \sum_i \vmathbb{1}_{\text{phase}_i} \cdot \|q - q^{\text{des}}_{\text{phase}_i}\| \)              & -0.2            \\
        \midrule
        Force         & Less Assi. Force     & \( \exp[-2\|F\| / \| \eta\|] \)                                                                      & 1.0             \\
        \bottomrule
    \end{tabular}
\end{table}

\subsubsection{Dancing}

To learn expressive whole-body motions, we utilize the dancing subset from the LAFAN1 motion dataset~\cite{harvey2020robust}, retargeted to the Unitree G1 robot by Unitree Robotics Company. A total of 8 motion clips are selected as expert demonstrations. These trajectories are interpolated to match the policy control frequency of 50\,Hz.

The task command \(c_t\) is defined as \([\Delta \mathbf{p}^{\text{cmd}}, \Delta \mathbf{q}^{\text{cmd}}, q^{\text{cmd}}]\), where \(\Delta \mathbf{p}^{\text{cmd}}\) denotes the desired displacement of the robot's base position, \(\Delta \mathbf{q}^{\text{cmd}}\) denotes the quaternion-based rotational difference of the base between adjacent timesteps, and \(q^{\text{cmd}}\) represents the desired joint positions. The policy is trained to follow these references using exponential tracking rewards based on the respective errors. The complete reward terms are listed in Table~\ref{tab:dance_rew}.

\begin{table}[H]
    \centering
    \caption{Reward Terms for Dancing Task.}
    \label{tab:dance_rew}
    \begin{tabular}{llll}
        \toprule
        \textbf{Type} & \textbf{Name}    & \textbf{Equation}                                                       & \textbf{Weight} \\
        \midrule
        \multirow{3}{*}{Task}
                      & Base Pos. Track  & \( \exp[-2500(\Delta \mathbf{p} - \Delta \mathbf{p}^{\text{cmd}})^2] \) & 10.0            \\
                      & Base Quat. Track & \( \exp[-100(\Delta \mathbf{q} - \Delta \mathbf{q}^{\text{cmd}})^2] \)  & 5.0             \\
                      & DOF Pos. Track   & \( \exp[-0.25(q - q^{\text{cmd}})^2] \)                                 & 20.0            \\

        \midrule
        \multirow{4}{*}{Reg.}
                      & Feet Slide       & \( \sum_i \vmathbb{1}_{\text{contact}, i} \cdot \|v_{foot, i}\| \)      & -1.0            \\
                      & Action Rate      & \( \|a_t - a_{t-1}\|^2 \)                                               & -0.01           \\
                      & DOF Acc.         & \( \|\ddot{q}\|^2 \)                                                    & -2.5e-7         \\
                      & DOF Torque       & \( \|\tau\|^2 \)                                                        & -1e-5           \\
        \midrule
        Force         & Less Assi. Force & \( \exp[-2\|F\| / \| \eta\|] \)                                         & 2.0             \\
        \bottomrule
    \end{tabular}
\end{table}

\section{EXPERIMENTAL RESULTS}
To validate the effectiveness and impact of the A2CF method, it is applied to three tasks: walking, backflip, and dancing. For both the walking and dancing tasks, real-world transfer experiments are conducted to verify the performance of the trained policies.

% \subsection{Implementation Details}

\begin{figure*}[htbp]
    \centering
    \includegraphics[width=1.0\linewidth]{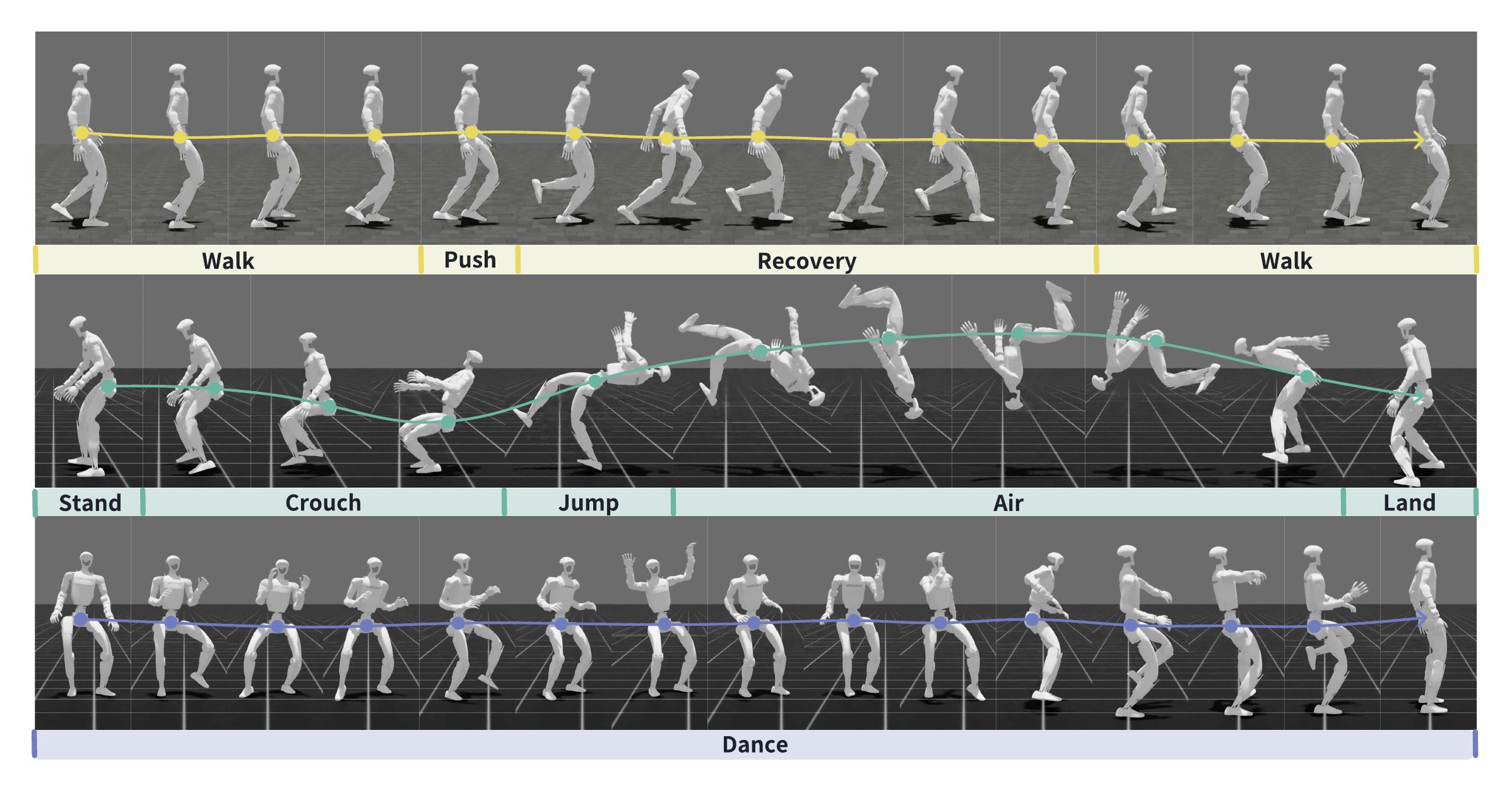}
    \caption{\textbf{Simulation of Walking, backflip, and Dancing Tasks.} The figure illustrates the execution of three tasks—walking, backflip, and dancing—in the simulation. The walking and backflip tasks have a 0.1\,s interval between frames, while the dancing task has a 1\,s interval. The walking sequence shows the robot recovering from a push disturbance. The backflip sequence highlights the robot's movement through five phases of the backflip.}
    \label{fig:sim_clip}
\end{figure*}

\subsection{Experimental Setup}\label{setup}

\subsubsection{Simulation Training}
Training is conducted on the Unitree G1 EDU humanoid robot, which consists of a total of 29 degrees of freedom (DoF): 7 DoF for each arm, 3 DoF for the waist, and 6 DoF for each leg. Isaac Lab \cite{mittal2023orbit} is employed as the simulation platform, and the Proximal Policy Optimization (PPO) algorithm \cite{schulman2017proximal, rudin2022learning} is used for policy optimization. The training is performed in parallel across 4096 environments on an NVIDIA GeForce 4090 GPU. The simulation time step is set to \( \Delta t = 0.001 \) s, and the environment step is set to \( \Delta t_{\text{env}} = 0.02 \) s, resulting in an execution frequency of 50 Hz for both the motion policy agent and the assistive force agent. For the walking and backflip tasks, each iteration takes approximately 5 seconds, while the dancing task, due to the need to sample expert data, requires approximately 8 seconds per iteration.

In the assistive force curriculum learning (Alg.~\ref{alg:curriculum}), the hyperparameters are set to $\epsilon = 0.5$ and $\delta = 0.2$. The initial distribution of the assistive force constraints in Section~\ref{ID} is shown in Table~\ref{tab:id}. In Section~\ref{RM}, the probability of random dropout for the mask is set to $\zeta = 0.2$.

\begin{table}[htbp]
    \centering
    \caption{Initial Distribution of Spatial Force}
    \label{tab:id}
    \begin{tabular}{llll}
        \toprule
        \textbf{Task} &         & \textbf{\( f_t \) Initial Bound} & \textbf{\( m_t \) Initial Bound} \\
        \midrule
        Walking       &         & \([40, 40, 10]\)                 & \([40, 40, 40]\)                 \\
        \midrule
        \multirow{5}{*}{Backflip}
                      & Phase 0 & \([40, 40, 10]\)                 & \([30, 30, 10]\)                 \\
                      & Phase 1 & \([40, 40, 10]\)                 & \([30, 30, 10]\)                 \\
                      & Phase 2 & \([40, 40, 100]\)                & \([30, 30, 30]\)                 \\
                      & Phase 3 & \([40, 40, 100]\)                & \([30, 30, 30]\)                 \\
                      & Phase 4 & \([40, 40, 10]\)                 & \([30, 30, 10]\)                 \\
        \midrule
        Dancing       &         & \([40, 40, 40]\)                 & \([40, 40, 40]\)                 \\
        \bottomrule
    \end{tabular}
\end{table}

\subsubsection{Real-World Deployment}

Upon completion of the simulation training, the learned policy is transferred to the real-world robot without additional fine-tuning for the Walking and Dancing tasks. The robot operates solely on the basis of proprioceptive sensors, which include joint angles, velocities, body orientation, and angular velocity, without the use of external sensor inputs. The control policy is executed at a frequency of 50 Hz on a personal computer equipped with an Intel Core i9-12900H CPU.

\subsection{Compared Methods}

To demonstrate the effectiveness of the proposed method in accelerating learning and achieving higher performance, the training curves of the Baseline and A2CF are compared. Additionally, a series of ablation experiments are conducted to validate the effectiveness of the PI, ID, and RM components, which are detailed in Section~\ref{PI}, Section~\ref{ID}, and Section~\ref{RM}, respectively. The methods compared are as follows:

\begin{enumerate}
    \item \textbf{Baseline}: The DreamWaQ framework implemented on the humanoid robot.
    \item \textbf{A2CF}: Our proposed method.
    \item \textbf{A2CF w/o PI}: A2CF without the use of privileged information (PI) for the assistive force agent.
    \item \textbf{A2CF w/o ID}: A2CF without the initial distribution of spatial force (ID). The spatial force is initialized with identical bounds across the \(x\), \(y\), and \(z\) dimensions.
    \item \textbf{A2CF w/o RM}: A2CF without the random mask (RM) mechanism for the assistive force.
\end{enumerate}

To ensure fairness, the reward function used in all methods, except for the assistive force-related terms, is kept consistent between A2CF and Baseline. Furthermore, all methods are trained with the same curriculum learning, domain randomization parameters, seed, and network architecture.

\subsection{Walking Task}
The walking policy is trained on rough terrain, which includes flat surfaces, slopes, rough terrains, discretized terrains, and stairs, organized into a 10-level difficulty curriculum. The highest stair height is 23 cm. Since the terrain level randomly shifts to any difficulty once the robot reaches its maximum level, the average maximum terrain level is approximately 6. The speed commands include a maximum forward velocity of 1.2 m/s, a maximum lateral velocity of 0.8 m/s, and a maximum angular velocity of 1.5 rad/s. Additionally, domain randomization is applied for friction coefficients, load, random velocity disturbances, and the position of the center of mass. The robot’s proficiency in the task is determined by whether it reaches the maximum terrain level, which then triggers the assistive force limit curriculum.

\begin{figure}[H]
    \centering
    \includegraphics[width=1.0\linewidth]{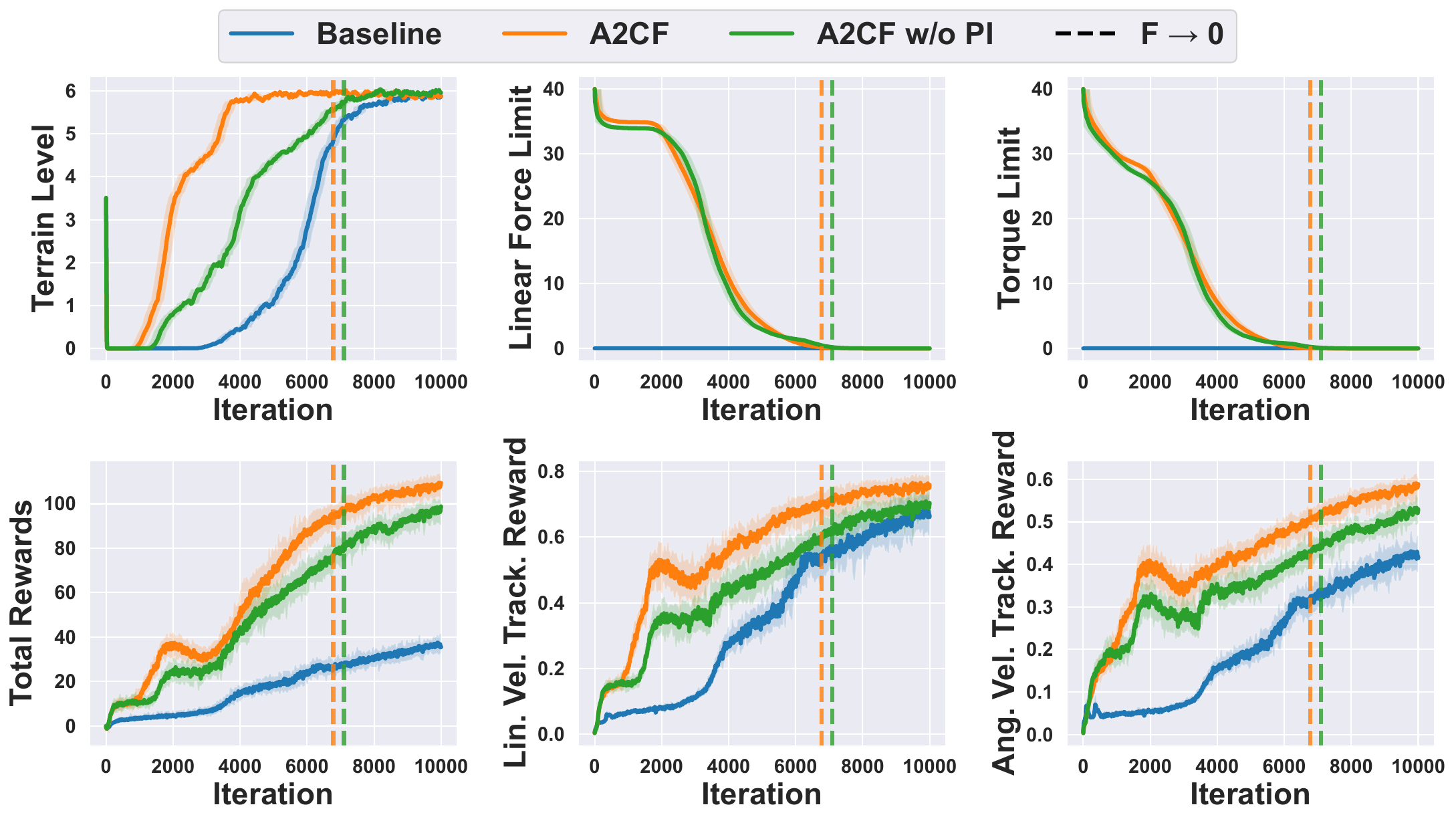}
    \caption{\textbf{Training Curves for the Walking Task.} The figure shows the terrain level, force limit curriculum, total rewards, and raw velocity tracking reward (ranging from 0 to 1). Vertical dashed lines indicate the point at which the assistive force becomes negligible.}
    \label{fig:walk_train}
\end{figure}

In this task, the A2CF algorithm is compared to the Baseline and A2CF without Privileged Information (w/o PI) during training. The results, shown in Figure~\ref{fig:walk_train}, demonstrate that A2CF learns walking capabilities much faster than the Baseline. Specifically, A2CF reaches the maximum terrain level around 4k iterations, while the Baseline requires approximately 10k iterations. Furthermore, the magnitude of the assistive force in A2CF falls below a negligible threshold of 0.1\,N—corresponding to an insignificant value for a 35\,kg robot—around 7k iterations, effectively enabling the emergence of an autonomous walking policy. Additionally, A2CF outperforms the Baseline in tracking both forward speed and angular velocity errors.

The ablation study on A2CF w/o PI highlights the critical role of privileged information, such as terrain perception, in enhancing the performance of the assistive force agent. The inclusion of PI accelerates the learning process and also causes the assistive force to decay to zero more rapidly. Additionally, the presence of PI results in improved velocity tracking performance, further demonstrating its beneficial impact on the overall control strategy.

\begin{figure}[htbp]
    \centering
    \includegraphics[width=1.0\linewidth]{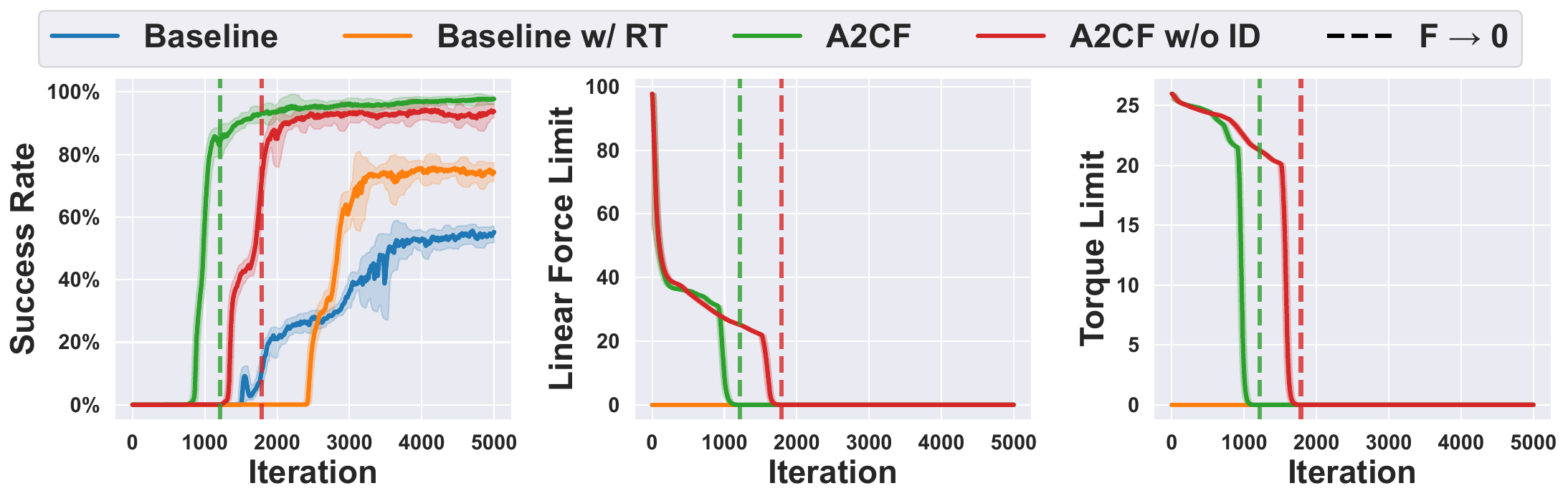}
    \caption{\textbf{Training Curves for the Backflip Task.} The figure shows the success rate of the backflip task during training and the corresponding assistive force limit curriculum. Vertical dashed lines indicate the point at which the assistive force becomes negligible.}
    % 写明阈值
    \label{fig:backflip_train}
\end{figure}

\subsection{Backflip Task}
The backflip policy is trained on flat terrain with randomized friction coefficients and payload masses. Task success is determined by whether the robot reaches the land phase and remains stable, which also triggers the assistive force limit curriculum.

In this task, the training success rates of Baseline, A2CF, and the ablated variant A2CF w/o ID are compared. Except for the additional assistive force reward, all other reward terms and weights remain consistent between Baseline and A2CF. Considering the reward sensitivity of the backflip task, an additional version of the Baseline is trained with manually tuned reward parameters, referred to as \textit{Baseline with Reward Tuning (Baseline w/ RT)}. The comparative results are shown in Figure~\ref{fig:backflip_train}.

Experimental results demonstrate that A2CF, benefiting from adaptive assistive force, learns the backflip policy more efficiently and achieves a success rate exceeding 90\%. The ablation study on A2CF w/o ID further validates the importance of a well-designed initial distribution of assistive force. In the backflip task, assistive force requirements vary significantly across different motion phases. Using uniform initial limits across all phases may reduce the diversity and effectiveness of the assistive force, ultimately hindering learning performance.

\begin{figure}[htbp]
    \centering
    \includegraphics[width=1.0\linewidth]{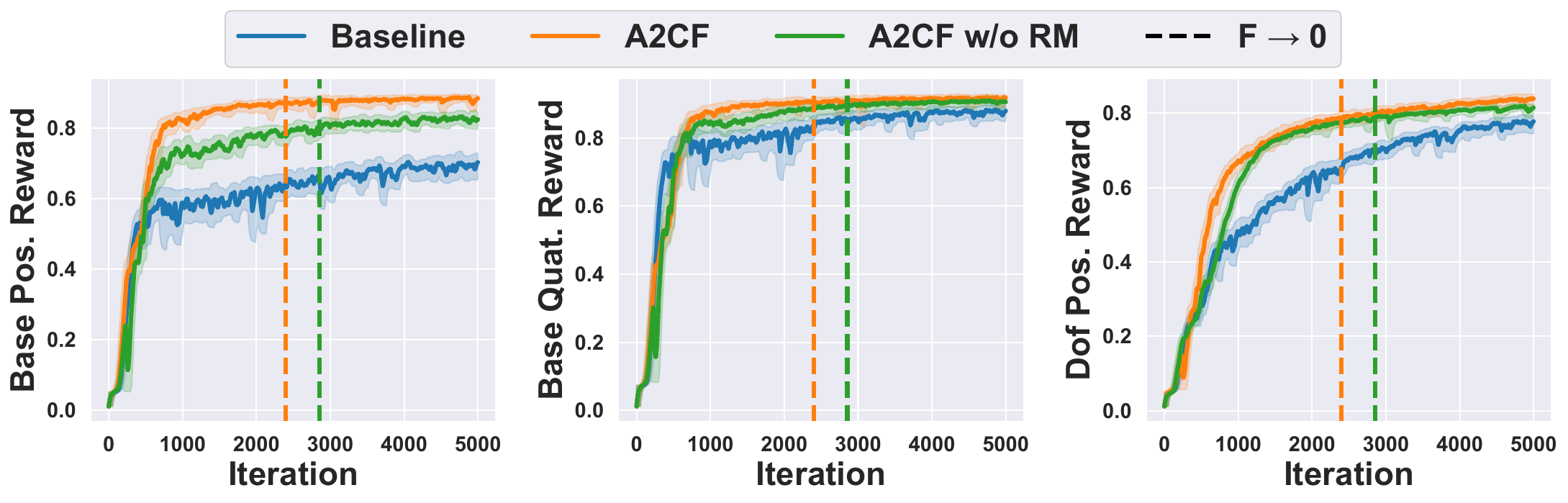}
    \caption{\textbf{Training Curves for the Dancing Task.} The figure shows the raw tracking rewards for base position, base orientation, and joint positions, without applying reward weights. Vertical dashed lines indicate the point at which the assistive force becomes negligible.}
    \label{fig:dance_train}
\end{figure}

\subsection{Dancing Task}
In the dancing task, training is conducted on flat terrain with randomized friction coefficients, payload masses, and center-of-mass positions. Task success is determined by whether the joint position tracking error reaches a threshold of 0.16, which triggers the assistive force limit curriculum.

In this task, the training reward curves for base position, base orientation, and joint position tracking are compared across Baseline, A2CF, and A2CF w/o RM. The experimental results, shown in Figure~\ref{fig:backflip_train}, indicate that A2CF achieves faster learning and better performance in joint position tracking accuracy compared to Baseline. While there is no significant improvement in the speed of base position and orientation tracking, A2CF demonstrates a substantial increase in tracking accuracy.

The ablation study on A2CF w/o RM highlights the importance of the random masking mechanism. By intermittently masking the assistive force, the method accelerates the agent's transition from reliance on assistive force to independent task execution, with assistive force decaying to zero earlier, enabling the agent to complete the task autonomously at an earlier stage.

\begin{figure}[htbp]
    \centering
    \includegraphics[width=1.0\linewidth]{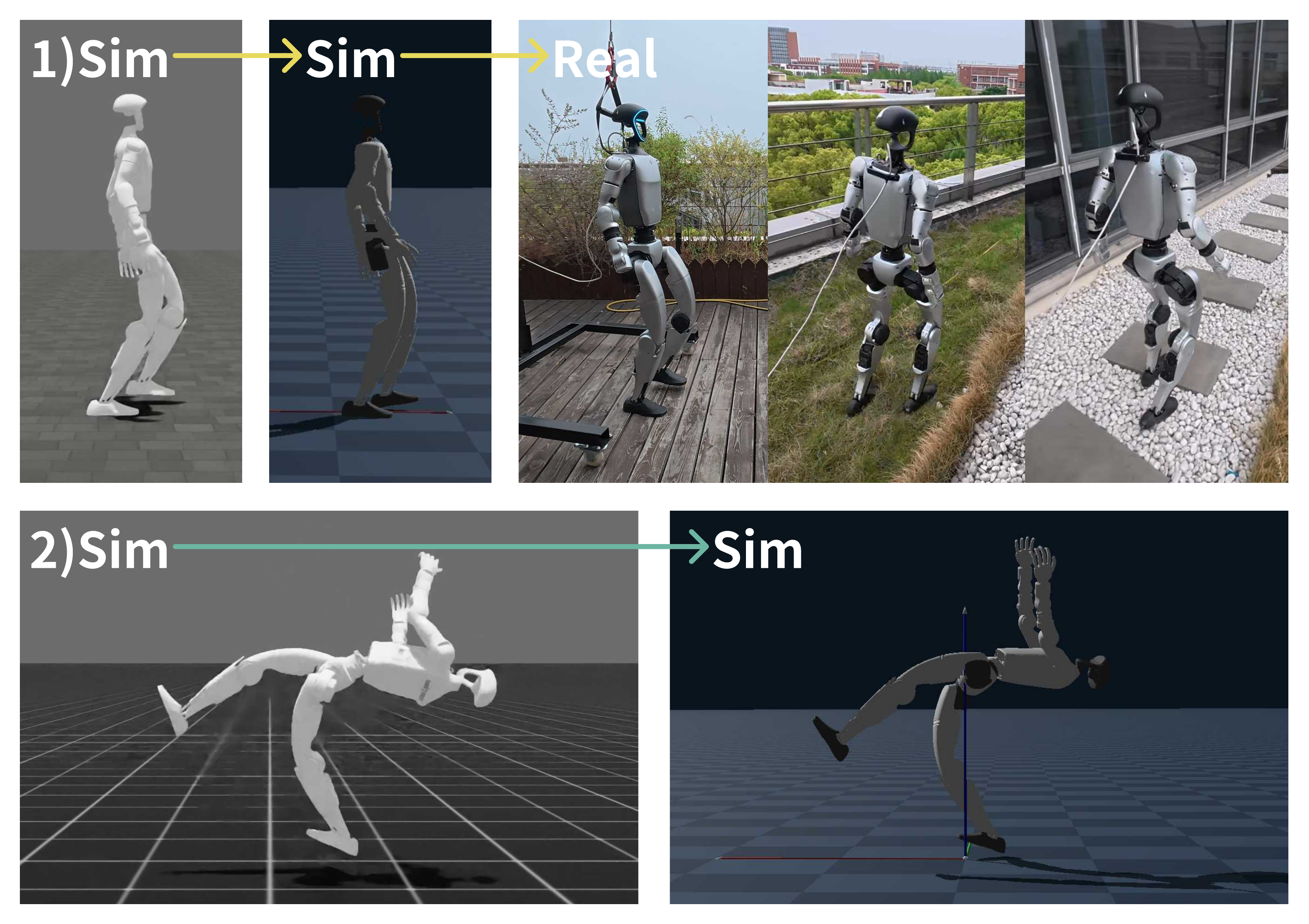}
    \caption{\textbf{Sim2Sim2Real Transfer Results for Walking and Sim2Sim Results for Backflip.} 1) The walking task shows Sim2Sim2Real transfer from IsaacLab (training domain) to Genesis (validation domain) and to the real-world environment (deployment domain). 2) The backflip task illustrates the Sim2Sim transfer from IsaacLab (training domain) to Genesis (validation domain).}
    \label{fig:walk_backflip_s2r}
\end{figure}

\subsection{Sim2Sim2Real Results}

To evaluate the effectiveness of the trained policies in real-world conditions, a Sim2Sim2Real transfer study was conducted. For the walking task, the policy was evaluated across diverse real-world terrains, including flat surfaces, grass, and uneven cobblestone paths, as demonstrated in the accompanying video. The results from these tests confirm that the walking policy, trained in simulation, generalizes well to real-world environments, as illustrated in Figure~\ref{fig:walk_backflip_s2r}. The robot successfully navigates different terrains, demonstrating robust performance despite environmental variations.

For the backflip task, real-world testing was not conducted due to limitations in funding and available space. However, the policy was transferred and validated in the Genesis simulator~\cite{Genesis}, where similar results were achieved. This validation process further supports the applicability of the trained policies in real-world scenarios. The Sim2Sim results for the backflip task are shown in Figure~\ref{fig:walk_backflip_s2r}.

\begin{figure}[htbp]
    \centering
    \includegraphics[width=1.0\linewidth]{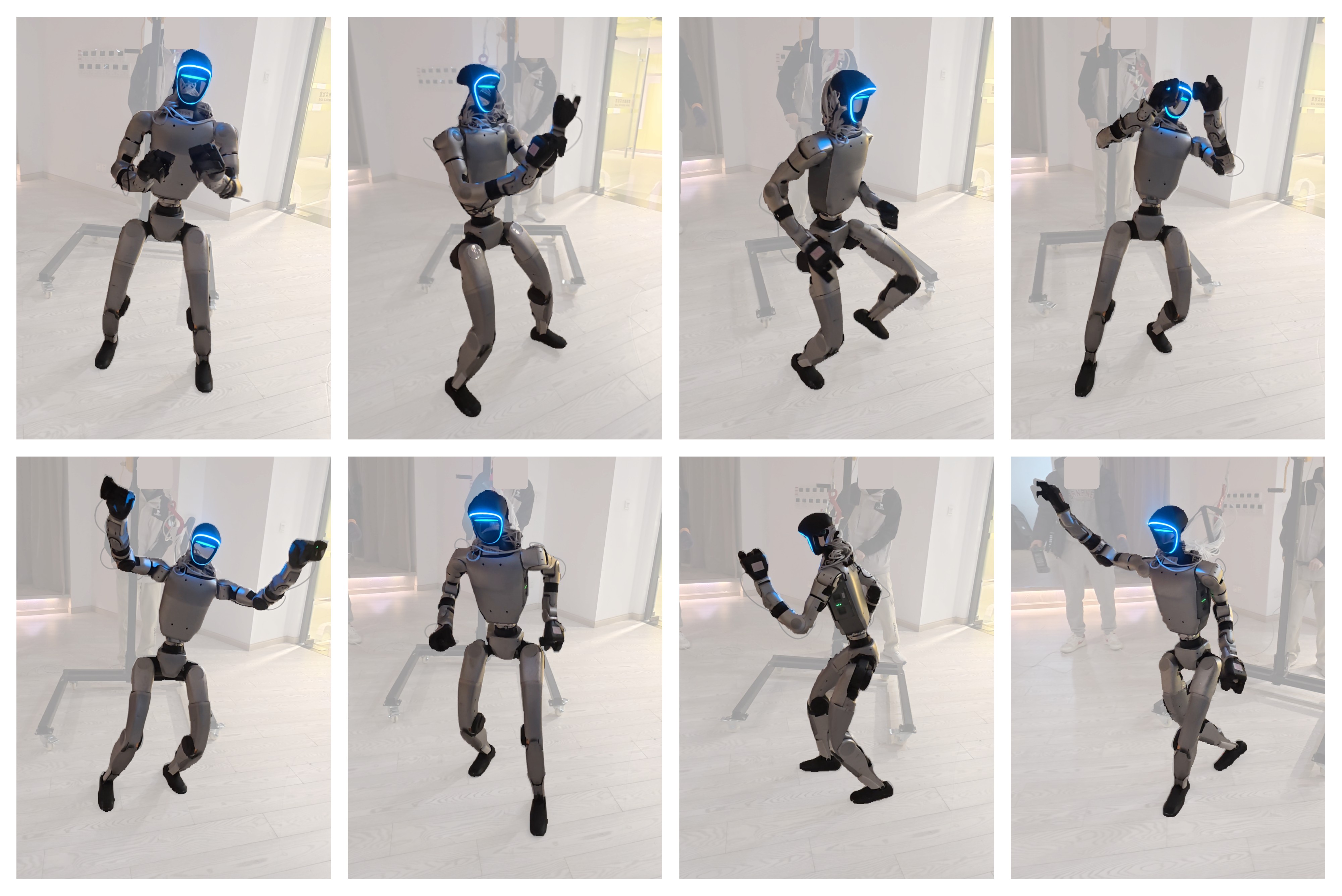}
    \caption{\textbf{Real-World Performance of the Dancing Task.} The figure shows the real-world deployment of the dancing policy, demonstrating the robot performing the dancing routine in a physical environment.}
    \label{fig:dance_real}
\end{figure}

Sim2Sim2Real transfer was also performed for the dancing task. The real-world results are presented in Figure~\ref{fig:dance_real}, where the robot performs the dance routine with high accuracy, demonstrating its ability to execute predefined motion sequences in a real-world environment.

The Sim2Sim2Real results for walking and dancing, along with the Sim2Sim results for the backflip, demonstrate the A2CF method’s ability to transfer learned policies across different domains. This capability is crucial for practical applications, where real-world conditions often differ from those in training simulations.

\section{CONCLUSIONS AND DISCUSSION}

In this work, we proposed the A2CF method to enhance learning efficiency and task performance in robotic tasks, including walking, backflip, and dancing. By incorporating adaptive assistive forces, privileged information, initial distribution, and random masking, A2CF accelerates the learning process and facilitates more autonomous task execution. Experimental results demonstrate that A2CF outperforms the Baseline in terms of learning speed and task success rates. Specifically, in the walking task, A2CF achieves faster learning and transitions to autonomy more efficiently. In the backflip task, A2CF achieves over 90\% success, while in the dancing task, it improves joint position tracking accuracy.

Currently, assistive forces are applied only to the base link of the robot. Future work could explore extending the application of assistive forces to other robot links, such as the hands, to further enhance performance. Additionally, incorporating objects in the environment, such as walls, could provide opportunities for the robot to autonomously explore and use these objects to gain assistive forces, introducing a new dimension of autonomy in real-world scenarios.

\addtolength{\textheight}{-12cm}   % This command serves to balance the column lengths

%%%%%%%%%%%%%%%%%%%%%%%%%%%%%%%%%%%%%%%%%%%%%%%%%%%%%%%%%%%%%%%%%%%%%%%%%%%%%%%%

%%%%%%%%%%%%%%%%%%%%%%%%%%%%%%%%%%%%%%%%%%%%%%%%%%%%%%%%%%%%%%%%%%%%%%%%%%%%%%%%



% \section*{ACKNOWLEDGMENT}

% The preferred spelling of the word acknowledgment in America is without an e after the g. Avoid the stilted expression, One of us (R. B. G.) thanks . . .  Instead, try R. B. G. thanks. Put sponsor acknowledgments in the unnumbered footnote on the first page.

%%%%%%%%%%%%%%%%%%%%%%%%%%%%%%%%%%%%%%%%%%%%%%%%%%%%%%%%%%%%%%%%%%%%%%%%%%%%%%%%

% References are important to the reader; therefore, each citation must be complete and correct. If at all possible, references should be commonly available publications.

\bibliographystyle{IEEEtran}
\bibliography{IEEEabrv, references}
% \bibliography{IEEEexample}

\end{document}